\documentclass{article}

\usepackage{microtype}
\usepackage{graphicx}
\usepackage{subfigure}
\usepackage{booktabs} 

\usepackage{hyperref}

\usepackage[accepted]{icml2024}

\usepackage{amsmath}
\usepackage{amssymb}
\usepackage{mathtools}
\usepackage{amsthm}

\usepackage[capitalize,noabbrev]{cleveref}

\theoremstyle{plain}

\theoremstyle{definition}

\theoremstyle{remark}

\begin{document}

\twocolumn[
\icmltitle{From Words to Worlds: Compositionality for Cognitive Architectures}




\begin{icmlauthorlist}
\icmlauthor{Ruchira Dhar}{yyy}
\icmlauthor{Anders Søgaard}{yyy}
\end{icmlauthorlist}
\icmlaffiliation{yyy}{Department of Computer Science, University of Copenhagen, Copenhagen, Denmark}
\icmlcorrespondingauthor{Ruchira Dhar}{rudh@di.ku.dk}

\icmlkeywords{Compositionality, Reasoning, LLM, Cognition}

\vskip 0.3in
]



\printAffiliationsAndNotice{}  


\begin{abstract}
Large language models (LLMs) are very performant connectionist systems, but do they exhibit more compositionality? More importantly, is that part of why they perform so well? We present empirical analyses across four LLM families (12 models) and three task categories, including a novel task introduced below. Our findings reveal a nuanced relationship in learning of compositional strategies by LLMs -- while scaling enhances compositional abilities, instruction tuning often has a reverse effect. Such disparity brings forth some open issues regarding the development and improvement of large language models in alignment with human cognitive capacities.
\end{abstract}

\section{Introduction}
\label{introduction}

Compositionality is a widely studied aspect of human cognition. \citet{fodor1988connectionism} claimed that non -- symbolic connectionist representations were inadequate for compositional understanding. The question turns on whether compositionality is acquired \cite{smolensky1987connectionist,chalmers1993connectionism}, or whether compositionality is merely a functional property \cite{van1990compositionality}. \citet{symons2014systematicity}, building on the arguments against connectionism in \cite{fodor1988connectionism}, argue that even if connectionist systems can stumble across an implementation of compositionality, this does not {\em explain} systematicity in their behaviour nor does it render them suitable cognitive architectures (see Appendix \ref{lab:appen1} for related work and Appendix \ref{lab:appen2} for further details). To serve as models of cognition or ``cognitive architectures", connectionist systems should ideally: 

\begin{itemize}
\item[i)] be compositional, i.e., have compositional representations and behaviour. 
\item[ii)] be compositional in a way that explains their behavior and performance, i.e., they should learn compositional strategies as a way to improve performance.
\end{itemize}

\begin{figure}[ht]
\vskip 0.2in
\begin{center}
\centerline{\includegraphics[width=\columnwidth]{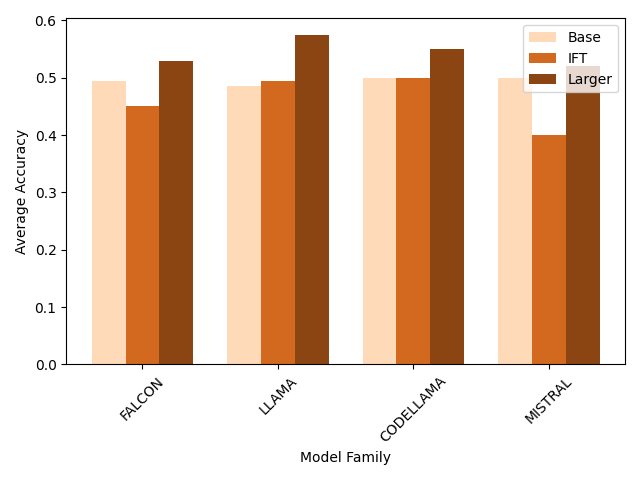}}
\caption{Model Accuracy trends for two setups (combined) with the ANTAILS Dataset.}
\label{icml-historical1}
\end{center}
\vskip -0.2in
\end{figure}

LLMs are now increasingly seen as possible models of human language \cite{mahowald2024dissociating,hu2024language} or cognition \cite{kauf2023event,hardy2023large,marjieh2023language,lamprinidis2023llm, aw2023instruction}, and it is therefore crucial to review the \citet{fodor1988connectionism} challenge from the perspective of LLMs. While there is work on measuring compositional abilities of LLMs \cite{dziri2024faith, li2024understanding, zhang2024can, wang2024can}, our focus is not to benchmark models for compositionality, but to examine its explanatory value in predicting performance and validating models as cognitive architectures.   

Scaling and instruction tuning are widely assumed to improve model alignment and generalization performance across a multitude of tasks ranging from natural language inference and textual entailment \cite{wei2022finetuned} to MMLU and BigBench \cite{longpre2023flan}  -- but are these improved performances a result of improved compositionality? Focusing on the domain of adjective -- noun (Adj -- N) composition, we propose three task types that can evaluate different aspects of compositional behaviour in LLMs and consider the impact of model size and instruction tuning on the compositional behaviour of such models. Finally, we discuss the importance of compositionality as a theoretical construct in validating connectionist cognitive architectures.

\section{Measuring Compositionality}

Over the years, several benchmarks have been developed to test compositionality of neural network models -- SCAN \cite{pmlr-v80-lake18a}, Lookup Table Composition \cite{DBLP:journals/corr/abs-1802-06467}, COGS \cite{kim-linzen-2020-cogs}, and PCFG Set \cite{hupkes2020compositionality}. However, we face a few issues when trying to leverage such datasets for testing today's LLMs- models pretrained on large amounts of text: 

\begin{itemize}
\item[a)] They are based on a  train -- test paradigm that is not easily applicable pretrained LLMs.
\item[b)] LLMs are trained on very large quantities of texts and may, as a consequence, have seen the test set expressions before.  
\item[c)]There is no congruence on what aspects of compositionality \cite{sun-etal-2023-validity} we test with these methods.
\end{itemize}

Some work considers compositional multi-hop reasoning \cite{lu2024chameleon, xu2024large,dziri2024faith,shao2022compositional}, but we focus on meaning construction from constituent representations. 

\begin{figure}[ht]
\vskip 0.2in
\begin{center}
\centerline{\includegraphics[width=\columnwidth]{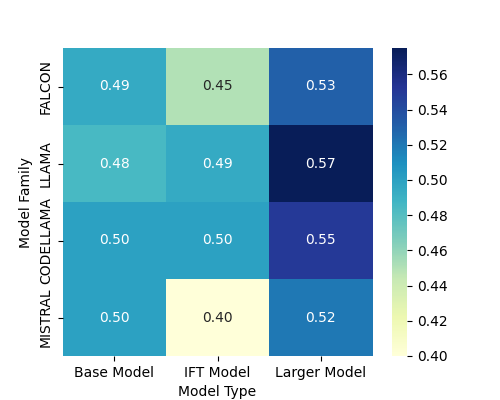}}
\caption{Heatmap for three model types and four model families on the ANTAILS dataset.}
\label{icml-historical2}
\end{center}
\vskip -0.2in
\end{figure}

We take inspiration from \citet{hupkes2020compositionality}'s tripartite distinction between aspects of compositionality, and introduce a test for each such aspect: 

\textbf{Substitutivity}: This involves the ability to understand the relatedness of words such that substituting a synonym in a complex expression should not be taken to alter the meaning of the complex expression. We test this with the \textbf{ANTAILS Dataset}, largely based on the AddOne Dataset \cite{pavlick-callison-burch-2016-babies}. For a given sentence with a noun (N) like \textit{The runner set a record}, we substitute N with an adjective -- noun combination like \textit{The runner set a new record} and test the model to see whether it can understand the entailment pattern. The model here has to maintain it's understanding of entailment patterns with adjective substitution.

\textbf{Systematicity \& Globalism}: This involves the ability to recombine known parts and rules and being able to productively use the parts in new contexts where constituents can have different behaviours \cite{carnap1988meaning}. We test this with the \textbf{PLANE Dataset} proposed by \citet{bertolini2022testing} that tests adjective -- noun entailment in a situation where the entailment pattern for an AN -- N and AN -- H (where AN is the adjective-noun combination, N is the noun and H is a hypernym of N) combination is already given and the model is tested on entailment of AN -- AH combination. This requires the model to employ systematicity (since the AN entailment pattern needs to be recombined in the AN -- AH statement) and also globalism ( since the entailment pattern of the AN -- AH combination needs to be inferred differently from the AN -- N and AN -- H combinations). 

\textbf{Over-generalization}: This involves the ability to distinguish between compositional and non -- compositional phenomena by measuring the distance of adjective -- noun combinations vs exocentric compounds. We test this with a new task type using a handcrafted toy dataset- the \textbf{COMPCOMB Dataset} -- which is a novel contribution of this work. Each data point consists of a triple -- a noun, an adjective that goes with the noun, and an exocentric compound which contains the noun. For example, (coat, trenchcoat and turncoat)-  when we take the word \textit{``coat"}, we know that \textit{``trenchcoat"} ( a special type of coat) is closely related to it but the exocentric compound \textit{``turncoat"} (a betrayer) is not since it is semantically different. This tests over -- generalization since the model needs to be able to distinguish between genuine compounds and combinations by avoiding generalization on the basis of surface forms.  

\section{Evaluating Models}

Our aim is to determine whether models' compositional abilities can indicate their trends of performance. There are two types of changes that have been shown to consistently impact the performance of models: 

\textbf{Scaling Parameters}: Research on LLM scaling laws -- \citet{kaplan2020scaling} and  \citet{hoffmann2022training} -- show that model performance for large language models get better with size i.e an increase in the number of parameters. 

\textbf{Instruction Tuning}: Several works \cite{wei2022finetuned,ouyang2022training,chung2024scaling} have shown the advantage of instruction finetuning (IFT) as a method to improve general performance of LLMs, especially for generalization to unseen tasks and alignment with human behaviour. 

Can these changes in performance of LLMs be explained by their compositional behaviour? To investigate this, we conduct analysis and evaluation across 4 families of models -- Falcon \cite{almazrouei2023falcon}, LLama \cite{touvron2023llama}, Codellama \cite{roziere2023code}, and Mistral \cite{jiang2023mistral}.

\subsection{ANTAILS Dataset}

For this dataset, we test 3 models for each model family -- the base model of 7B (Base), an instruction tuned version of the same (IFT), and a larger model -- with two different kinds of setups:  one involving a two -- choice question scenario where we determine accuracy by fixed rank precision (P@k) to evaluate the model output (Setup 1) and another in which we use the log probabilities of the model for two completions (entails vs does not entail) as an indication of the model's judgement (Setup 2). Furthermore, for both setups we include two prompt variations for the evaluation and the result table shows the average accuracy for each model across the prompt variations.

\textbf{Results}: We observe that for all families of models, the Larger Model always performs better than the Base Model (Figure 1\& 2). However, the impact of instruction tuning is inconsistent with performance decreasing for the two models, remaining the same for Codellama, and increasing for Llama.

\begin{table}[t]
\caption{ANTAILS Experiment in Setup 1}
\label{tab:antailsexperim1}
\vskip 0.15in
\begin{center}
\begin{small}
\begin{sc}
\resizebox{\columnwidth}{!}{
\begin{tabular}{lccc}
\toprule
\textbf{Model Family} & \textbf{Base Model} & \textbf{IFT Model} & \textbf{Larger Model} \\
\midrule
FALCON     & 0.50$\pm$0.01 & 0.46$\pm$0.01 & 0.54$\pm$0.03 \\
LLAMA 2      & 0.50$\pm$0.01 & 0.54$\pm$0.04 & 0.60$\pm$0.01 \\
CODELLAMA  & 0.50$\pm$0.01 & 0.50$\pm$0.01 & 0.55$\pm$0.02 \\
MISTRAL    & 0.50$\pm$0.01 & 0.30$\pm$0.04 & 0.51$\pm$0.02 \\
\bottomrule
\end{tabular}
}

\end{sc}
\end{small}
\end{center}
\vskip -0.1in
\end{table}

\begin{table}[t]
\caption{ANTAILS Experiment in Setup 2}
\label{tab:example2}
\vskip 0.15in
\begin{center}
\begin{small}
\begin{sc}
\resizebox{\columnwidth}{!}{
\begin{tabular}{lccc}
\toprule
\textbf{Model Family} & \textbf{Base Model} & \textbf{IFT Model} & \textbf{Larger Model} \\
\midrule
FALCON     & 0.49$\pm$0.01 & 0.44$\pm$0.02 & 0.52$\pm$0.02 \\
LLAMA 2      & 0.47$\pm$0.03 & 0.45$\pm$0.05 & 0.55$\pm$0.05 \\
CODELLAMA  & 0.50$\pm$0.01 & 0.50$\pm$0.01 & 0.55$\pm$0.02 \\
MISTRAL    & 0.50$\pm$0.03 & 0.50$\pm$0.20 & 0.53$\pm$0.07 \\
\bottomrule
\end{tabular}
}

\end{sc}
\end{small}
\end{center}
\vskip -0.1in
\end{table}

\subsection{PLANE Dataset}

\begin{figure}[ht]
\vskip 0.2in
\begin{center}
\centerline{\includegraphics[width=\columnwidth]{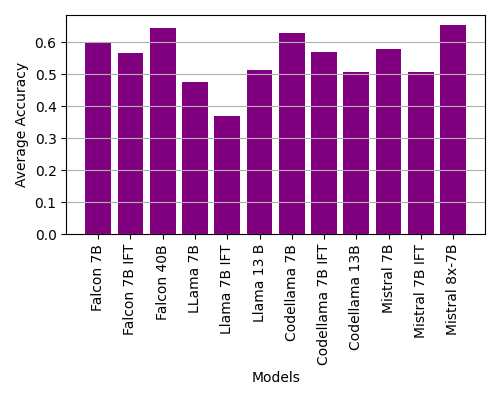}}
\caption{Model accuracy trends for PLANE dataset.}
\label{icml-historical3}
\end{center}
\vskip -0.2in
\end{figure}
 
For this dataset, we also have a setup that is exactly similar to the previous one. Since the dataset is divided by types of adjectives, we also present the results classified by the different adjectival categories. 

\textbf{Results}: Similar to the ANTAILS dataset, we observe in Figure 3 that the overall model performance, within a model family, across two setups improves with size and worsens with instruction tuning (Tables 3 and 4). However, in the case of within -- family comparison in the Codellama family of models,  the larger model (13B) is worse than the base (7B) indicating that training a general LM with code and scaling it might not always have positive impacts on compositional reasoning. Similar trends were also observed by \citet{ma2024at}, where introduction of code at pretraining stage gives worse performance in logical reasoning tasks.

Figure 4 shows the comparative analysis of results across different adjective classes -- I ( Intersective), N (Subsective), and O ( Intensional). Most models perform worse for subsective adjectives. Interestingly, \citet{redolfi2024processing} notes that children also acquire subsectives the slowest during the period of language acquisition.

\begin{figure}[ht]
\vskip 0.2in
\begin{center}
\centerline{\includegraphics[width=\columnwidth]{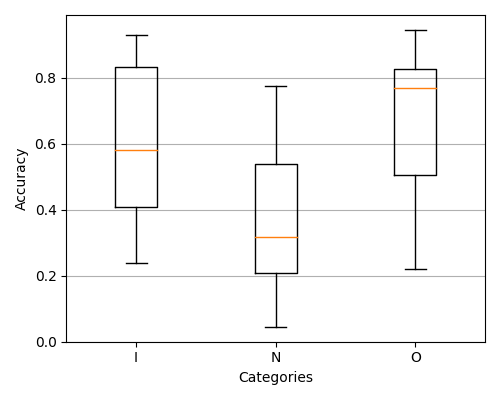}}
\caption{Average accuracies of models across 3 classes of adjectives.}
\label{icml-historical4}
\end{center}
\vskip -0.2in
\end{figure}

\begin{table}[t]
\caption{PLANE Experiments in Setup 1}
\label{tab:experiments1}
\vskip 0.15in
\begin{center}
\begin{small}
\begin{sc}
\resizebox{\columnwidth}{!}{
\begin{tabular}{lccc}
\toprule
\textbf{Models} & \multicolumn{3}{c}{\textbf{Accuracies}} \\
\cmidrule(lr){2-4}
 & \textbf{I} & \textbf{N} & \textbf{O} \\
\midrule
falcon-7B        & 0.95$\pm$0.01   & 0.07$\pm$0.07 & 0.69$\pm$0.13 \\
falcon-7B-instruct    & 0.70$\pm$0.16   & 0.03$\pm$0.02 & 0.85$\pm$0.05 \\
falcon-40B       & 0.80$\pm$0.16   & 0.14$\pm$0.1  & 0.98$\pm$0.02 \\
Llama-2-7B-hf         & 0.03$\pm$0.03   & 0.98$\pm$0.02 & 0.01$\pm$0.01 \\
Llama-2-7B-chat-hf     &0.25$\pm$0.05   & 0.53$\pm$0.50 & 0.12$\pm$0.12 \\
Llama-2-13B-hf       & 0.04$\pm$0.03   & 0.95$\pm$0.05   & 0.14$\pm$0.07 \\
Codellama-7B-hf     & 0.78$\pm$0.02   & 0.25$\pm$0.04 & 0.81$\pm$0.01 \\
Codellama-7B-Instruct-hf & 0.55$\pm$0.08   & 0.35$\pm$0.05 & 0.60$\pm$0.15 \\
Codellama-13B-hf    & 0.18$\pm$0.01   & 0.90$\pm$0.01   & 0.03$\pm$0.01 \\
Mistral-7B-v0.1      & 0.05$\pm$0.01   & 0.80$\pm$0.15 & 0.60$\pm$0.02 \\
Mistral-7B-Instruct-v0.1   & 0.64$\pm$0.01   & 0             & 0.50$\pm$0.05 \\
Mixtral-8x7B-v0.1    & 0.98$\pm$0.03   & 0.04$\pm$0.02 & 0.98$\pm$0.02 \\
\bottomrule
\end{tabular}
}
\end{sc}
\end{small}
\end{center}
\vskip -0.1in
\end{table}

\begin{table}[t]
\caption{PLANE Experiments in Setup 2}
\label{tab:experiments2}
\vskip 0.15in
\begin{center}
\begin{small}
\begin{sc}
\resizebox{\columnwidth}{!}{
\begin{tabular}{lccc}
\toprule
\textbf{Models} & \multicolumn{3}{c}{\textbf{Accuracies}} \\
\cmidrule(lr){2-4}
 & \textbf{I} & \textbf{N} & \textbf{O} \\
\midrule
falcon-7B        & 0.91$\pm$0.05   & 0.07$\pm$0.03 & 0.90$\pm$0.09 \\
falcon-7B-instruct    & 0.97$\pm$0.02   & 0.06$\pm$0.01 & 0.80$\pm$0.12 \\
falcon-40B       & 0.87$\pm$0.06   & 0.17$\pm$0.10 & 0.91$\pm$0.05 \\
Llama-2-7B-hf         & 0.45$\pm$0.06   & 0.51$\pm$0.14 & 0.88$\pm$0.01 \\
Llama-2-7B-chat-hf     & 0.55$\pm$0.07   & 0.45$\pm$0.11 & 0.32$\pm$0.02 \\
Llama-2-13B-hf       & 0.89$\pm$0.01   & 0.15$\pm$0.02 & 0.91$\pm$0.05 \\
Codellama-7B-hf     & 0.88$\pm$0.02   & 0.20$\pm$0.01 & 0.86$\pm$0.11 \\
Codellama-7B-Instruct-hf & 0.65$\pm$0.07   & 0.39$\pm$0.06 & 0.89$\pm$0.10 \\
Codellama-13B-hf    & 0.47$\pm$0.08   & 0.65$\pm$0.05 & 0.82$\pm$0.05 \\
Mistral-7B-v0.1       & 0.77$\pm$0.21   & 0.27$\pm$0.21 & 0.98$\pm$0.02 \\
Mistral-7B-Instruct-v0.1   & 0.48$\pm$0.02   & 0.50$\pm$0.09 & 0.92$\pm$0.02 \\
Mixtral-8x7B-v0.1    & 0.58$\pm$0.10   & 0.49$\pm$0.15 & 0.85$\pm$0.15 \\
\bottomrule
\end{tabular}
}
\end{sc}
\end{small}
\end{center}
\vskip -0.1in
\end{table}

\begin{figure}[ht]
\vskip 0.2in
\begin{center}
\centerline{\includegraphics[width=\columnwidth]{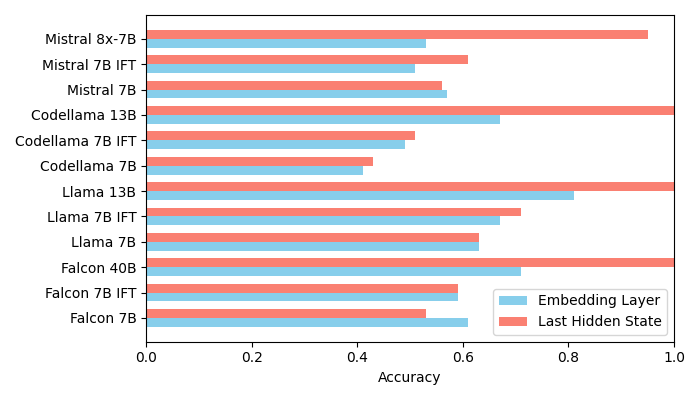}}
\caption{Model Accuracy trends for two setups with the COMPCOMB Dataset }
\label{icml-historical}
\end{center}
\vskip -0.2in
\end{figure}

\begin{table}[t]
\caption{COMPCOMB Experiments for two setups}
\label{tab:experimentscomp2}
\vskip 0.15in
\begin{center}
\begin{small}
\begin{sc}
\resizebox{\columnwidth}{!}{
\begin{tabular}{lcc}
\toprule
\textbf{Models} & \textbf{Accuracy (EL)} & \textbf{Accuracy (LHS)} \\
\midrule
falcon-7B       & 0.61                     & 0.53                   \\
falcon-7B-instruct    & 0.59                     & 0.59                   \\
falcon-40B       & 0.71                     & 1                      \\
Llama-2-7B-hf         & 0.63                     & 0.63                   \\
Llama-2-7B-chat-hf     & 0.67                     & 0.71                  \\
Llama-2-13B-hf        & 0.81                     & 1                      \\
Codellama-7B-hf     & 0.41                     & 0.43                   \\
Codellama-7B-Instruct-hf & 0.49                     & 0.51                   \\
Codellama-13B-hf    & 0.67                     & 1                      \\
Mistral-7B-v0.1       & 0.57                     & 0.56                   \\
Mistral-7B-Instruct-v0.1   & 0.51                     & 0.61                   \\
Mixtral-8x7B-v0.1    & 0.53                     & 0.95                   \\
\bottomrule
\end{tabular}
}
\end{sc}
\end{small}
\end{center}
\vskip -0.1in
\end{table}

\subsection{COMPCOMB Dataset}

For each datapoint this dataset, we evaluate the accuracy of model in terms of comparative cosine distance analysis of the embeddings/hidden states of models. If the dist(N, AN) < dist (N,H) we consider the model accurate since it is able to capture the semantic similarity of N -- AN as compared to N --H. We do this for two types of embedding for each model -- for Setup 1, we use the initial embeddings from the embedding layer (EL) while in Setup 2, we access the last hidden state of the model (LHS).

\textbf{Results}: The accuracy of models across all families increases with size (Table 5). In Figure 5, we notice that while the embedding layer still shows over -- generalization for larger models, the last hidden state representation has much better performance. For instruction tuned models, the performance of the embedding layer varies. 

Refer to appendices \ref{lab:appenmodels}, \ref{lab:appendata} \& \ref{lab:appentask} for additional details on models, datasets and task setups.

\section{Conclusion}

Cognitive architectures should arguably be performant and exhibit compositionality, and the induction of compositional strategies should be explanatory of their performance. LLMs, as candidate cognitive architectures, are clearly performant, behave compositionally (as seen by their performance on the ANTAILS and PLANE datasets), and their representations appear compositional (as seen by their performance on the COMPCOMB dataset). However, when it comes to how explanatory the induction of compositional strategies are of performance improvements, we observe different patterns for different LLMs:
 \begin{itemize}
     \item [1)] Scaling models improves their generalization capabilities \cite{hendrycks2020pretrained, desai2020calibration} and overall performance \cite{kaplan2020scaling, hoffmann2022training}. Compositional behaviour also improves with scaling across model families. This could indicate that the induction of compositional strategies is explanatory of improvements with scaling.
     \item [2)]Instructing finetuning has been shown to improve alignment and result in performance gains across several task types \cite{wei2022finetuned,ouyang2022training}. However, we see that compositional performance does not always improve with instruction tuning. Performance gains from instruction tuning do not correlate with improved compositional behaviour.
 \end{itemize}

In sum, while scaling often leads to more compositional models, instruction tuning does not show similar trends. Recent work \cite{ghosh2024closer} has shown that instruction tuning sometimes degrades performance. Our results indicate that one source of error may be reduced compositionality. Performance is multi -- faceted, and compositionality may be explanatory of some performance gains, not others. If we think cognitive architectures should learn compositional strategies \cite{fodor1988connectionism, symons2014systematicity}, {\em and} that LLMs could potentially be cognitive architectures \cite{lamprinidis2023llm, sumers2023cognitive, zhao2023more}, we must evaluate if the compositionality of LLMs is explanatory of their performance and be precise about what (relevant) performance is at play. This work is, to the best of our knowledge, the first step in that direction.

\section*{Limitations}

The focus on adjective-noun combinations in tasks might provide a limited view of the models' overall compositional abilities. Broader investigation across various domains is necessary to understand models' capabilities, limitations, and behavior trends in scaled versus instruction -- tuned models. Additionally, incorporating error analysis and interpretability techniques will uncover underlying mechanisms and biases in model outputs, guiding improvements and ensuring more transparent interpretations and application of results. We plan on incorporating such changes in future iterations of this work.

\section*{Impact Statement}

This paper presents work whose goal is to advance the field of 
machine learning. There are many potential societal consequences 
of our work, none of which we feel must be specifically highlighted here.

\bibliography{example_paper}
\bibliographystyle{icml2024}


\newpage
\clearpage
\appendix

\section{Related Work}
\label{lab:appen1}

Most of the earlier work on testing compositionality in connectionist systems was centered around two main types: 

1. Testing compositional abilities: Most works \cite{pmlr-v80-lake18a,DBLP:journals/corr/abs-1802-06467,kim-linzen-2020-cogs,hupkes2020compositionality} have a training and testing paradigm where models were considered to be performing compositional generalization if they were able to successfully handle unseen test sequences. 

2. Enhancing compositional abilities: This area of research was focused on what enhancements to connectionist models -- architectures, training methods, or data -- could provide improved compositionality. Some like \citet{socher2010learning} involved combining syntactic parse trees with connectionist architectures to learn compositional functions, allowing models to be ‘compositional by design' while other work like \citet{lake2023human} proposed a novel method for training neural networks via a series of compositional tasks that endows them with systematic generalization capabilities.

Recent work has shifted the focus to testing compositional generalization in pretrained models via tasks that require no further training. There is some research that focuses on prompting to enable better results in compositional tasks \cite{drozdov2023compositional,chen2024skillsincontext}. 

However, much of recent work \cite{li2024understanding, alabdulakreem2024securellm, shao2023compositional, zhang2024can} interprets compositionality to be multi -- hop reasoning which is not ``true" compositionality which was originally discussed as a feature of human language and cognition \cite{frege1892sense, fodor1988connectionism}.

\section{The Compositionality Debate- Symbolism vs Connectionism}
\label{lab:appen2}

The concept of compositionality has a long history in linguistic and cognitive science -- it was perhaps first discussed in detail by \citet{frege1892sense} in the context of how natural language expressions were assigned meanings. \citet{partee1995lexical} formulated the so-called {\em principle of compositionality}:

\begin{quote}The meaning of a complex expression is determined by the meanings of its constituent parts and the rules used to combine them.\end{quote}

Compositionality has long been considered a cornerstone of human cognitive capabilities and was notably discussed in \citet{fodor1988connectionism} as the reason for the systematicity of human thought -- how the ability to think a thought is linked to the ability to also have related thoughts. Non -- symbolic connectionist representations were, in the view of Fodor and Pylyshyn, inadequate and unviable. Instead, they claimed that: (i) only classical or symbolic representations can give rise to compositional and, in turn, systematic behaviour; and (ii) neural networks do not have classical representations and thus they cannot exhibit compositional understanding or behaviour.

A central tenet of \citet{fodor1988connectionism} was also not how connectionist systems could not behave compositionally but why they could not serve as viable cognitive architectures. The then highly debated topic of acquisition of English past tense is discussed to point out that even though \citet{plunkett1999connectionist} finds a way to show that connectionist systems can simulate this pattern, \cite{pinker1988language} is correct in asserting that it does not in any way {\em explain} the actual cognitive process. \citet{symons2014systematicity} further elucidate the point claiming that even if connectionist systems can stumble across an implementation of compositionality, this does not {\em explain} systematicity in their behaviour nor does it render them suitable cognitive architectures.

\citet{smolensky1987connectionist} and \citet{chalmers1993connectionism} were instrumental in challenging the prevailing skepticism towards connectionist networks by asserting that these networks have the capacity to embody classical representations through their intricate connection weights and activation patterns, thereby exhibiting compositional behavior. This assertion stems directly from the inherent expressivity of connectionist models, which allow them to capture complex relationships and hierarchies within data.

\citet{van1990compositionality} suggested that neural networks displayed {\em functional compositionality} instead of the traditional `concatenative' compositionality discussed by Fodor. According to Van Gelder, neural networks demonstrate a form of compositionality where the functions computed by individual neurons or layers combine in a compositional manner to produce complex behaviors, without necessarily relying on explicit concatenation of discrete symbols.

\citet{cummins1996representations} argued against the dichotomy between classical and non-classical representations, contending that the distinction fails to hold ground since classical representations themselves can exhibit non -- compositional characteristics. This perspective underscores the complexity and fluidity inherent in the nature of representations, suggesting that compositionality is not necessarily tied to a specific type of representation but rather emerges from the interactions and transformations within a system.

\begin{figure}
    \centering
    \includegraphics[width=0.99\linewidth]{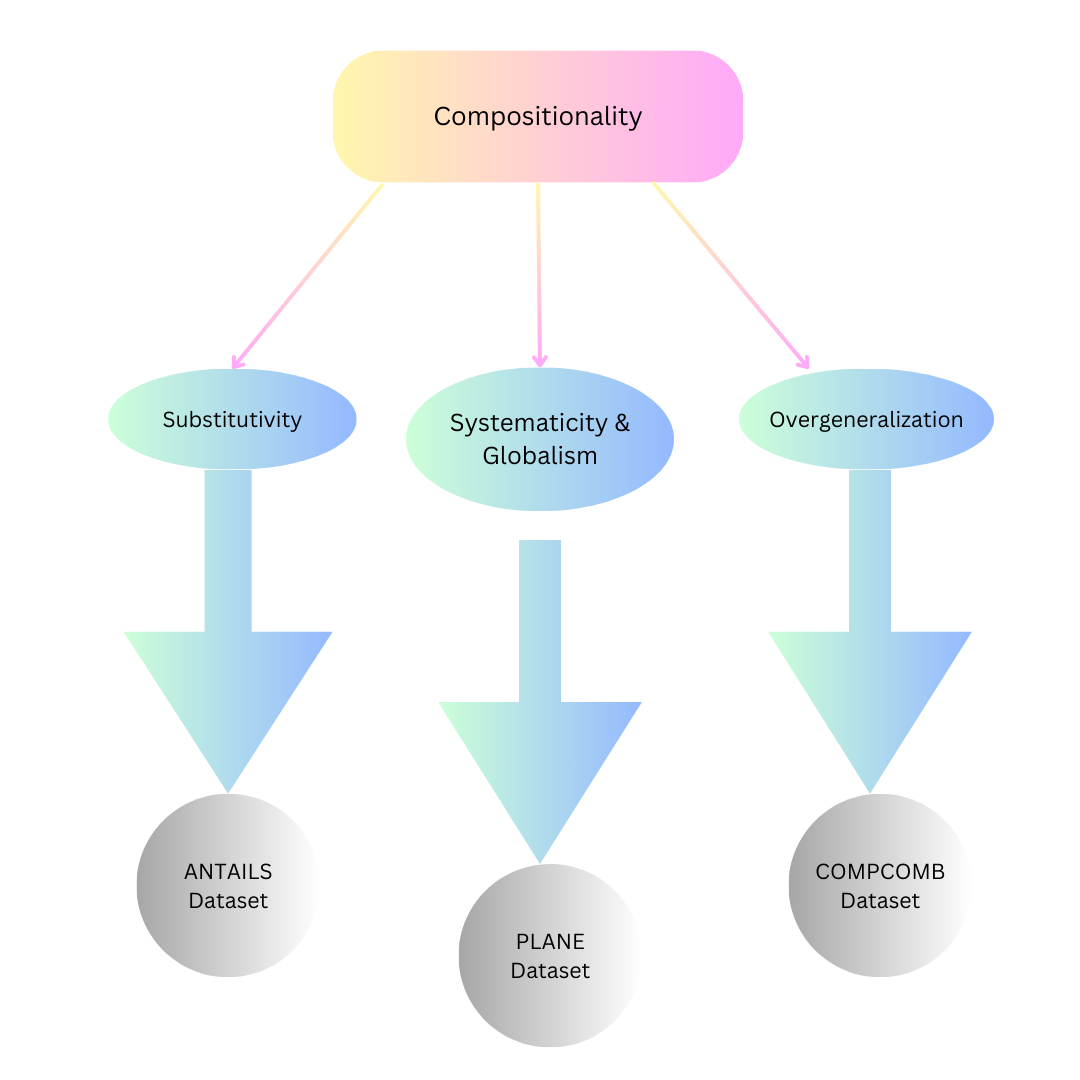}
    \caption{Testing Compositionality: Experimental Setup for LLMs }
\end{figure}

\section{Models}
\label{lab:appenmodels}

The models used here are all based on the transformer architecture but are decoder -- only models. For each model family, we use 3 variants: 

1. Falcon Family: falcon-7B, falcon-7B-instruct, falcon-40B

2. Llama 2 Family: Llama-2-7B-hf, Llama-2-7B-chat-hf, Llama-2-13B-hf

3. Codellama Family : Codellama-7B-hf, Codellama-7B-Instruct-hf, Codellama-13B-hf

4. Mistral Family: Mistral-7B-v0.1, Mistral-7B-Instruct-v0.1, Mixtral-8x7B-v0.1

We provide a summary of models used and their Huggingface Hub links in Table 6 to enable easy reproduction and use.

\begin{table}[t]
\caption{Models used and corresponding Huggingface Hub Links}
\label{tab:models}
\vskip 0.15in
\begin{center}
\begin{small}
\begin{sc}
\resizebox{\columnwidth}{!}{
\begin{tabular}{lc}
\toprule
\textbf{Model Name} & \textbf{Model Link} \\
\midrule
falcon-7B & \url{https://huggingface.co/tiiuae/falcon-7b} \\
falcon-7B-instruct & \url{https://huggingface.co/tiiuae/falcon-7b-instruct} \\
falcon-40B & \url{https://huggingface.co/tiiuae/falcon-40b} \\
Llama-2-7B-hf & \url{https://huggingface.co/meta-llama/Llama-2-7b-hf} \\
Llama-2-7B-chat-hf & \url{https://huggingface.co/meta-llama/Llama-2-7b-chat-hf} \\
Llama-2-13B-hf & \url{https://huggingface.co/meta-llama/Llama-2-13b-hf} \\
Codellama-7B-hf & \url{https://huggingface.co/meta-llama/CodeLlama-7b-hf} \\
Codellama-7B-Instruct-hf & \url{https://huggingface.co/meta-llama/CodeLlama-7b-Instruct-hf} \\
Codellama-13B-hf & \url{https://huggingface.co/meta-llama/CodeLlama-13b-hf} \\
Mistral-7B-v0.1 & \url{https://huggingface.co/mistralai/Mistral-7B-v0.1} \\
Mistral-7B-Instruct-v0.1 & \url{https://huggingface.co/mistralai/Mistral-7B-Instruct-v0.1} \\
Mixtral-8x7B-v0.1 & \url{https://huggingface.co/mistralai/Mixtral-8x7B-v0.1} \\
\bottomrule
\end{tabular}
}
\end{sc}
\end{small}
\end{center}
\vskip -0.1in
\end{table}

\section{Datasets}
\label{lab:appendata}

We measure different aspects of compositionality with 3 task types/ datasets: 

1. ANTAILS: It is the adjective noun entailment dataset. The dataset is influenced by \citet{pavlick-callison-burch-2016-babies} but we found certain discrepancies in the dataset due to which we slightly modify and build our own dataset as shown in Figure 7.

2. PLANE: It adjective -- noun hypernym inference pattern testing introduced in \citet{bertolini2022testing} and we use the same dataset as shown in Figure 8. 

3. COMPCOMB: This is a task (as shown in Figure 9) that centers around measuring the distance of compounds vs adjective -- noun combinations in the embedding space of models. It is a novel contribution of this work.  

\begin{figure}
    \centering
    \includegraphics[width=0.99\linewidth]{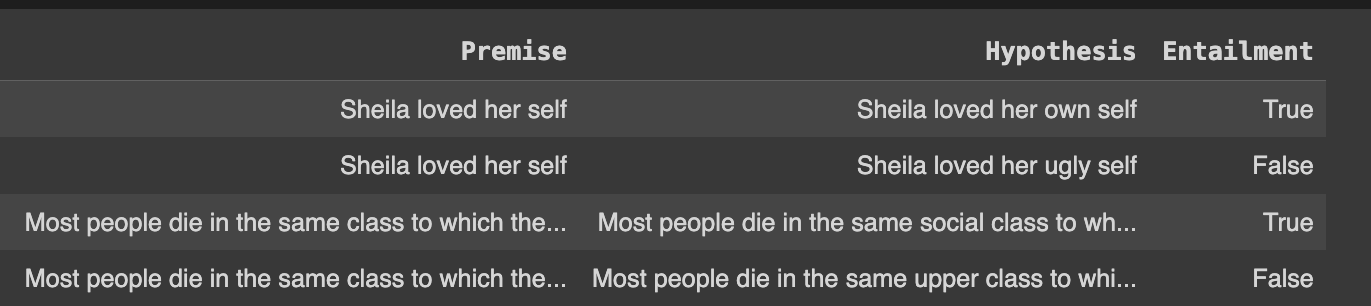}
    \caption{ANTAILS Dataset}
\end{figure}

\begin{figure}
    \centering
    \includegraphics[width=0.99\linewidth]{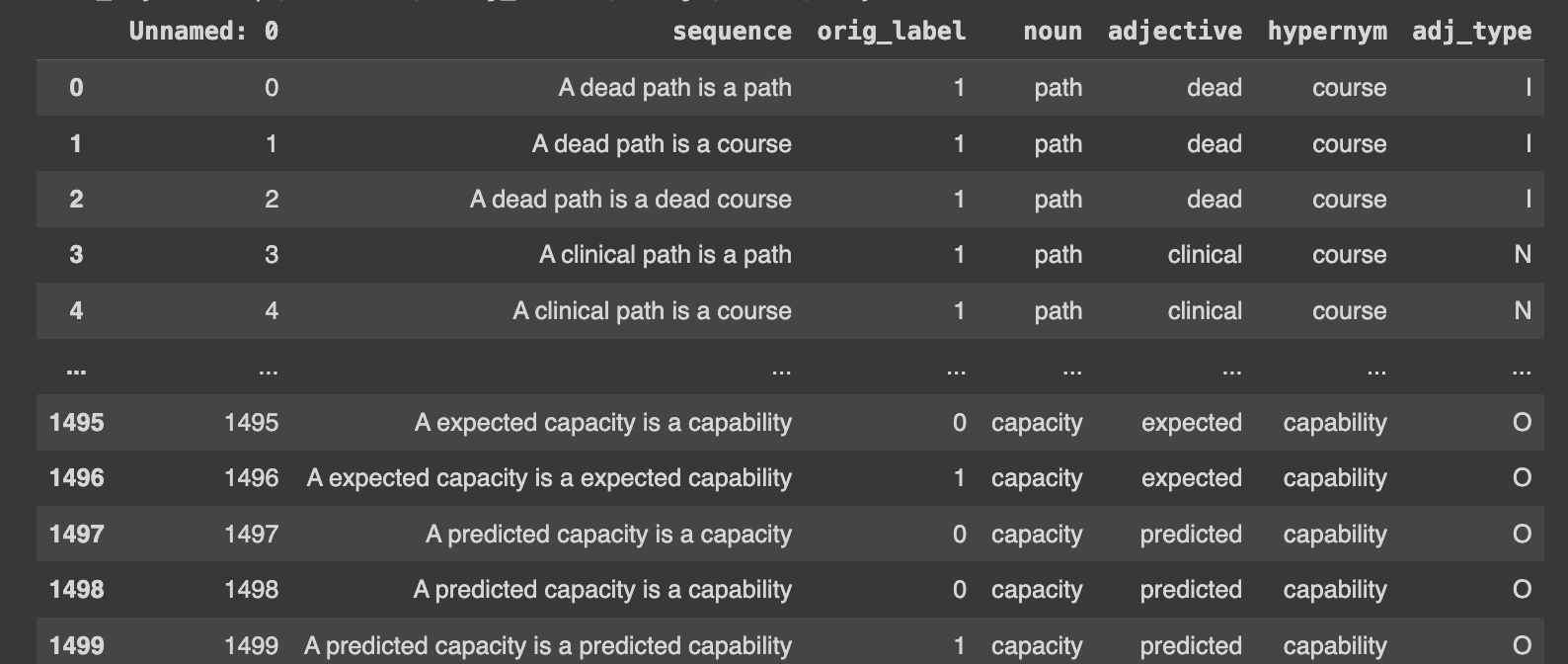}
    \caption{PLANE Dataset}
\end{figure}

\begin{figure}
    \centering
    \includegraphics[width=0.7\linewidth]{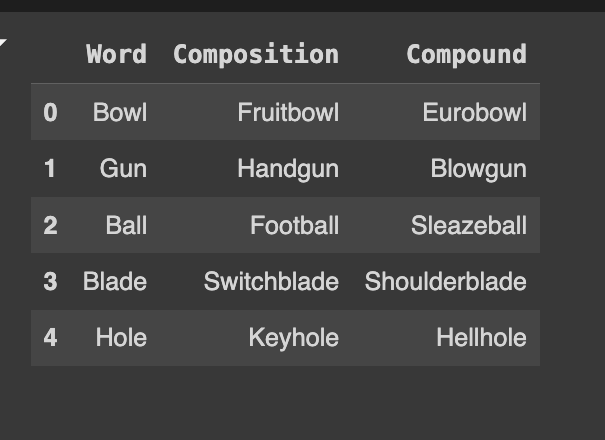}
    \caption{COMPCOMB Dataset}
\end{figure}

\section{Task Setup}
\label{lab:appentask}

To investigate the trends of learning compositional strategies, we investigate two types of models -- base models and instruction models -- and use similar prompts for all models. Some motivations for our prompting choice setup are the following:

1. For all models, we do a zero -- shot prompt setting to attempt an unbiased comparison of general vs instruction tuned models. Works on instruction tuning \cite{wei2022finetuned,sanh2021multitask} indicate that such models have good zero -- shot task performance and thus we chose this prompting mode for all models to try and avoid undue bias.

2. We wanted to use a similar prompt structure across models in our work to maintain uniformity of evaluation. Since using instruction format prompts would disadvantage a non instruction tuned model and research indicates instruction tuning improves general reasoning and performance, we chose to avoid specific prompting methods involving advanced instructions. Non -- instruction prompts can effectively serve as robust evaluation tools, helping to assess the model's true understanding and generalization ability beyond the training data \cite{peng2023instruction, sun2023evaluating}.

For the ANTAILS and PLANE datasets, we use two task setups: 

1. \textbf{Two -- Choice QA}: The first setup gives models statements indicating entailment and non entailment as two options and the model choice of option is considered. We avoid using the yes-no setup to prevent possible yes -- bias outputs.

2. \textbf{Logprob Calculation}: The second setup involves passing in the prompt with dataset samples and calculating the log probabilities of the model for a statement indicating entailment and one indicating non -- entailment. The statement assigned higher completion log probability is considered to be the model output. 

For both setups, we use two prompts and average the outputs to calculate our results. We observe similar trends across different prompt choices.

For the COMPCOMB dataset, the above task settings do not apply since we directly compare representations of the model from the embedding layer and the last hidden layer. 

\end{document}